\documentclass[sigconf]{acmart}

\usepackage{booktabs} 
\usepackage{overpic}
\usepackage{rotating}

\setcopyright{rightsretained}

\acmDOI{10.475/123_4}

\acmISBN{123-4567-24-567/08/06}

\acmConference[MM'18]{ACM Multimedia conference}{October 2018}{Seoul, Korea}
\acmYear{2018}
\copyrightyear{2018}

\acmPrice{15.00}

\acmSubmissionID{163-A12-B3}

\def\eg{\emph{e.g.,~}}
\def\etal{{\em et al.~}}

\graphicspath{{Imgs/}}
\begin{document}
\title{Face Sketch Synthesis Style Similarity: \\ A New Structure Co-occurrence Texture Measure}


\author{
Deng-Ping Fan$^1$,
ShengChuan Zhang$^2$,
Yu-Huan Wu$^1$,\\
Ming-Ming Cheng$^{1\dag}$,
Bo Ren$^{1}$,
Rongrong Ji$^2$  and
Paul L Rosin$^3$
}
\authornote{
$^1$ CCCE, Nankai University \quad\quad
$^2$ Xiamen University \quad\quad
$^3$ Cardiff University\\
$\dag$ Ming-Ming Cheng (cmm@nankai.edu.cn) is the corresponding author.\\
Source Code: \url{http://dpfan.net/Scoot/}
}

\newcommand{\sArt}{state-of-the-art}
\newcommand{\figref}[1]{Fig. \ref{#1}}
\newcommand{\secref}[1]{Sec. \ref{#1}}
\newcommand{\tabref}[1]{Tab.~\ref{#1}}
\newcommand{\eqnref}[1]{(Eq. \ref{#1})}

\newcommand{\fdp}[1]{{\textcolor{red}{#1}}}
\newcommand{\wyh}[1]{{\textcolor{blue}{#1}}}

\begin{abstract}
    Existing face sketch synthesis (FSS) similarity measures are sensitive to slight
    \emph{image degradation} (\eg noise, blur).
    However, human perception of the similarity of two sketches will
    consider both structure and texture as essential factors and is not
    sensitive to slight (``pixel-level'') mismatches. 
    %
    Consequently, the use of existing similarity measures can lead to better algorithms receiving a lower score 
    than worse algorithms.
    This unreliable evaluation has significantly hindered the development of the FSS field.
    To solve this problem, we propose a
    novel and robust style similarity measure called \textbf{Scoot-measure} (Structure
    CO-Occurrence Texture Measure), which simultaneously evaluates ``block-level''
    spatial structure and co-occurrence texture statistics.
    In addition, we further propose 4 new meta-measures and create 2 new datasets 
    to perform a comprehensive evaluation of several widely-used FSS measures on 
    two large databases.
    Experimental results demonstrate that our measure not only provides a reliable evaluation
    but also achieves significantly improved performance. 
    Specifically, the study indicated a higher degree (78.8\%) of correlation between 
    our measure and human judgment than the best prior measure (58.6\%).
    %
    Our code will be made available.
\end{abstract}

%
%
%


\maketitle

\section{Introduction}\label{sec:Introduction}

\begin{figure}[t!]
\centering
    \begin{overpic}[width=\columnwidth]{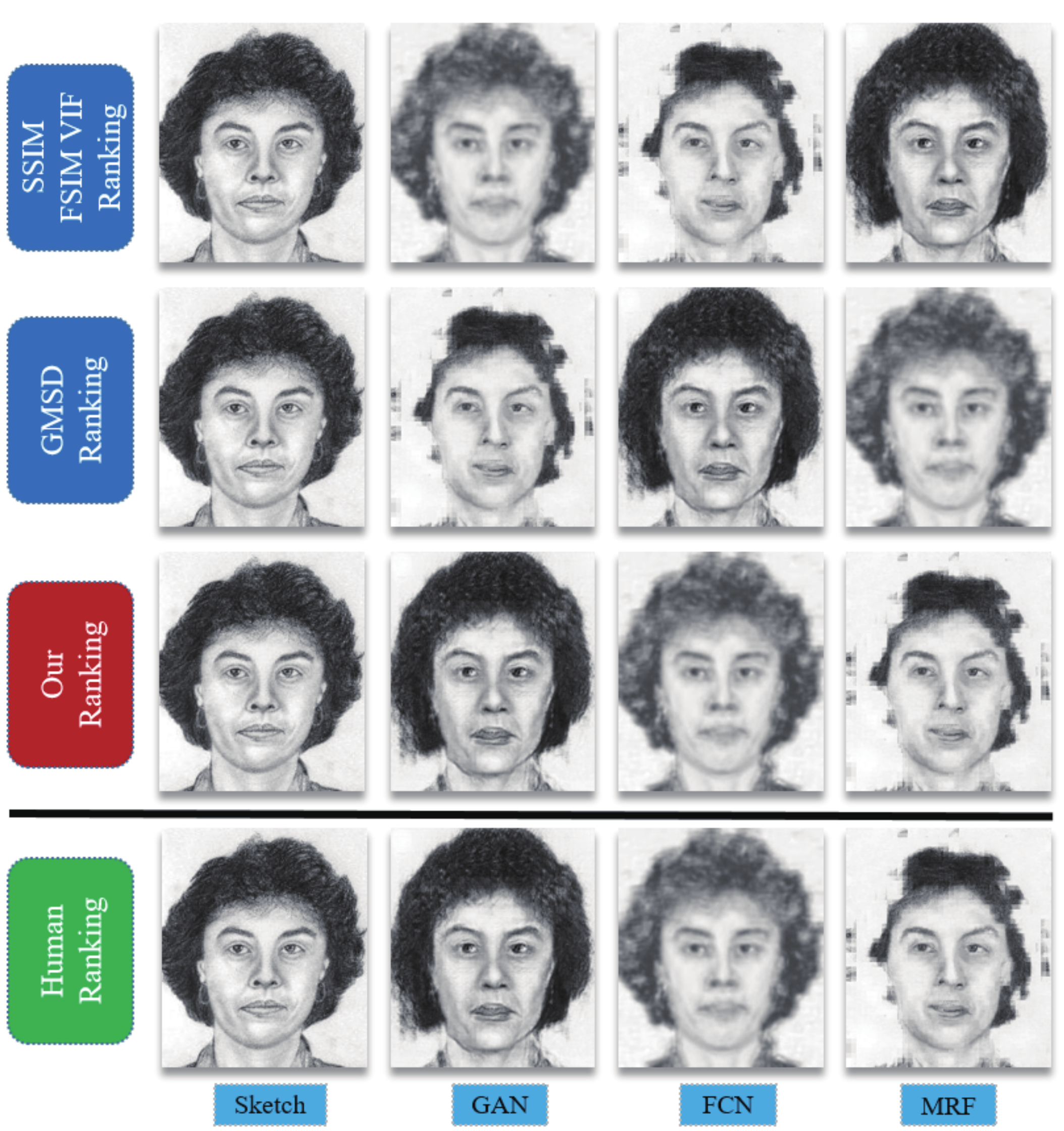}
    \put(82,100){$3^{rd}$}
    \put(62,100){$2^{nd}$}
    \put(42,100){$1^{st}$}
    \end{overpic}
    \vspace{-20pt}
    \caption{Inaccuracy of current evaluation measures.
    We compare the ranking of face sketches synthesized by 3 face sketch 
    synthesis (FSS) algorithms: GAN~\cite{isola2017image}, FCN~\cite{zhang2015end},
    and MRF~\cite{wang2009face}. According to the human ranking (last row), the GAN ranks first, 
    followed by
    the FCN and MRF sketches. The GAN synthesizes the face of
    both structure and texture most similarly, with respect to the
    sketch drawn by the artist. The FCN captures the structure but lost lots of
    textures. The MRF almost completely destroyed the structure of
    the face. However, the most common measures (SSIM~\cite{wang2004image}, 
    FSIM~\cite{zhang2011fsim}, VIF~\cite{sheikh2006image}, and GMSD~\cite{xue2014gradient})
    fail to rank the sketches correctly. Only our measure (third row)
    correctly ranked the results.}\label{fig:FirstExample}
\end{figure}

Please take a look at \figref{fig:FirstExample} in which you can see several synthesized face sketches (GAN~\cite{isola2017image}, FCN~\cite{zhang2015end}, MRF~\cite{wang2009face}). 
Which one do you think is closer to the ground-truth (GT) sketch drawn by the artist?
While this comparison task seems trivial for humans, to date the most common
measures (\eg FSIM~\cite{zhang2011fsim}, SSIM~\cite{wang2004image},
VIF~\cite{sheikh2006image}, and GMSD~\cite{xue2014gradient}) cannot reliably rank the algorithms according to their similarity to humans.
This is the problem that we address in this paper, and to solve it we propose a novel
measure that does much better than existing measures in terms of ranking algorithm
syntheses compared to the GT sketch.

Face sketch synthesis (FSS) has been gaining popularity in
the design community (\eg face artistic styles synthesis)~\cite{berger2013style,wang2014comprehensive}
and is being used for digital
entertainment~\cite{gao2008local,yu2012semisupervised,zhang2017compositional}, and multimedia
surveillance law enforcement (\eg sketch based face recognition)
~\cite{tang2004face,best2014unconstrained}. In such applications, the comparison of a synthesized
sketch against a GT sketch is crucial in evaluating the
quality of a face sketch synthesis algorithm.

\begin{figure*}[t!]
\centering
    \begin{overpic}[width=.8\textwidth]{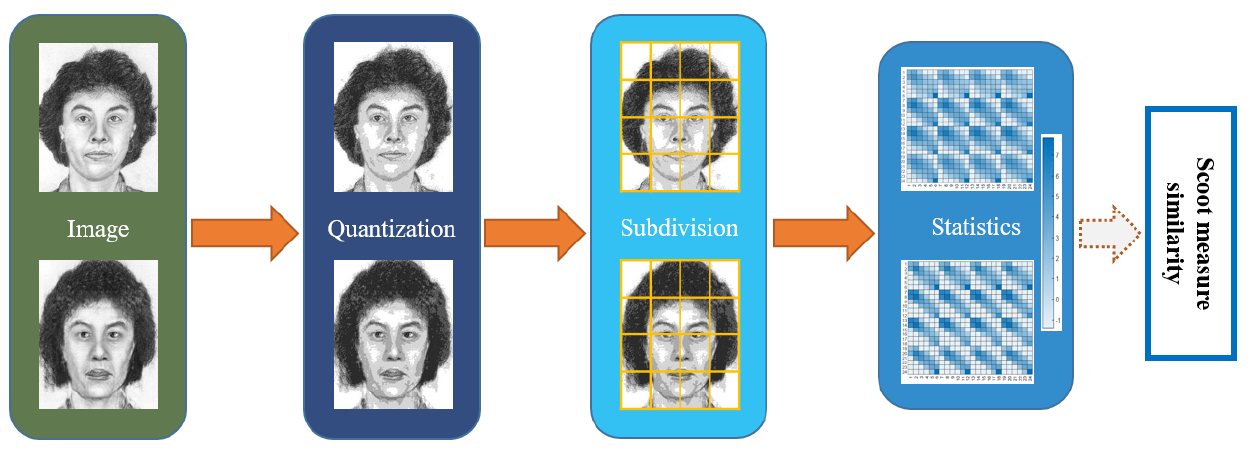}
    \put(-10,26){GT Sketch:}
    \put(-10,8){Synthesis:}
    \end{overpic}
    \caption{The Scoot-measure framework for FSS style similarity evaluation between 
    GT sketch and synthesis. We first quantize the two input sketches 
    to a set of grades. Then, we divide them into blocks to consider their spatial 
    structure. Thirdly, we extract their ``block-level'' co-occurrence texture statistics. Finally, 
    the style similarity can be computed.
    }\label{fig:framework}
\end{figure*}

To date, the most widely-used~\cite{wang2018random,wang2017data,peng2016multiple,zhang2017face,zhang2015face,zhang2016robust,wang2017unified,zhang2016fast,wang2014comprehensive}
evaluation measures are that of Structural SIMilarity
(SSIM)~\cite{wang2004image}, Visual Information Fidelity
(VIF)~\cite{sheikh2006image}, Feature SIMilarity
(FSIM)~\cite{zhang2011fsim}, and Gradient Magnitude Similarity Deviation (GMSD)~\cite{xue2014gradient}.
These measures were originally designed for {\em pixel-level} Image Quality
Assessment (IQA) which aims to detect types of image degradation such as
Gaussian blur, jpeg, and jpeg 2000 compression. Therefore these measures should
be sensitive to slight (\eg pixel-level) change in image.
Psychophysics~\cite{zucker1989two} and prior works
(\eg style for lines of drawing)~\cite{grabli2004programmable,zucker2005computation,freeman2003learning}
indicate that style is an essential factor in sketches.
Note that human perception of the style similarity of two sketches will
consider both of structure and texture~\cite{wang2016evaluation} as the essential factor
rather than being sensitive to ``pixel-level'' mismatches.
As a consequence, current FSS measures often provide unreliable evaluation due to the different nature of their task.
Consequently, a better algorithm may receive a lower score than a worse algorithm (see \figref{fig:FirstExample}). 
This unreliable evaluation has significantly hindered the development of the FSS field. 
However, designing a reliable style similarity measure is difficult.
Ideally, such a measure should:
\begin{itemize}
  \item be \textbf{insensitive to slight mismatches}
  since real-world sket-ches drawn by artists do not exactly match each pixel to the 
  original photos
  (\secref{sec:Insensitive}, \secref{sec:rotation});
  \vspace{5pt}

  \item hold \textbf{good holistic-content capture capability}, and should for example
  score a complete state-of-the-art (SOTA) sketch higher than the results of only preserving
  light strokes (\secref{sec:Content});
  \vspace{5pt}

  \item obtain a \textbf{high correlation} to human perception so that
  the sketch can be directly used in various subjective
  applications, especially for law enforcement and entertainment
  (\secref{sec:Judgment}).
\end{itemize}
As far as we know, no previous measures can meet all these goals
simultaneously and provide reliable evaluation results.

To this end, we propose a new and robust style similarity measure called
\textbf{Scoot-measure} (Structure CO-Occurrence Texture Measure),
which simultaneously evaluates ``block-level'' spatial structure and
co-occurrence texture statistics. We experimentally demonstrate that
our measure offers a reliable evaluation and achieves significantly
improved performance. Specially, the experiment indicated a higher degree
(78.8\%) of correlation between our measure and the human judgment than
the best prior measure (58.6\%). Our main contributions are:

1) We propose a simple and robust style similarity measure for face sketch synthesis that
provides a unified evaluation capturing both structure and texture.
We experimentally show that our measure is significantly better than most of the current widely used measures and some alternatives including GLRLM~\cite{galloway1974texture},
Gabor~\cite{gabor1946theory}, Canny~\cite{canny1987computational}, and Sobel~\cite{sobel1990isotropic}.

2) To assess the quality of style similarity measures, we further propose 4 new meta-measures 
based on 3 hypotheses and build two new human ranked face sketches datasets.
These two datasets contain 676, and 1888 synthesized face sketches, respectively.
We use the two datasets to examine the ranking consistency between current
measures and human judgment.

\section{Related Works}
As discussed in the introduction, a reliable similarity 
evaluation measure in FSS does not exist~\cite{wang2016evaluation}. 
Previous researchers just used the standard measures in IQA to evaluate the similarity of FSS. 

Wang \etal~\cite{wang2004image} proposed the \textbf{SSIM} index.
Since the structural information reflects the object structure features in the scenes,
SSIM computes the structure similarity, the luminance and contrast comparison using a sliding window on the local patch.

Sheikh and Bovik~\cite{sheikh2006image} proposed the \textbf{VIF} measure which evaluates the
image quality by quantifying two kinds of information.
One is obtained via the human visual system (HVS) channel, with the input
ground truth and the output reference image information. The other is
obtained via the distortion channel, called distortion information.
And the result is the ratio of these two kinds of information.

Physiological and psychological studies found that the perceived
features of human vision show great consistency with the phase
consistency of Fourier series at different frequencies.
Therefore, Zhang \etal~\cite{zhang2011fsim} choose the phase congruency
as the primary feature. Since phase congruency is not affected by the contrast,
which is a key point of image quality, the image gradient magnitude
is chosen as the second feature. Then they proposed a low-level feature
similarity measure called \textbf{FSIM}.

Recently, Xue~\etal~\cite{xue2014gradient} devised a simple measure named
gradient magnitude similarity deviation (\textbf{GMSD}), where the pixel-wise
gradient magnitude similarity is utilized to obtain image local quality,
and the standard deviation of the overall GMS map is calculated as the
final image quality index. Their measure achieves the SOTA performance
compared with the other measures in IQA.

However, all the above-mentioned measures were originally designed for IQA which
focuses on pixel-level image degradation, and so they are sensitive to slight
resizing (see \figref{fig:MM1}) of the image. Although some of them consider structure
similarity, the structure is based on pixel-wise rather than pair-wise measurements.
By observing sketches we found that the pair-wise co-occurrence texture is more
suitable to capture the style similarity (also see our discussion in \secref{sec:discussion}) than the pixel-wise texture.

\section{Scoot measure}\label{sec:methodology}
In this section, we introduce our novel measure to evaluate the 
style similarity for FSS. The important property of our measure is that
it captures the pair-wise \textbf{co-occurrence texture} statistics in the
``block-level'' \textbf{spatial structure}. As a result, our Scoot-measure
works better than the current widely-used measures. The framework is shown in \figref{fig:framework}.

\begin{figure}[t]
\centering
    \begin{overpic}[width=\columnwidth]{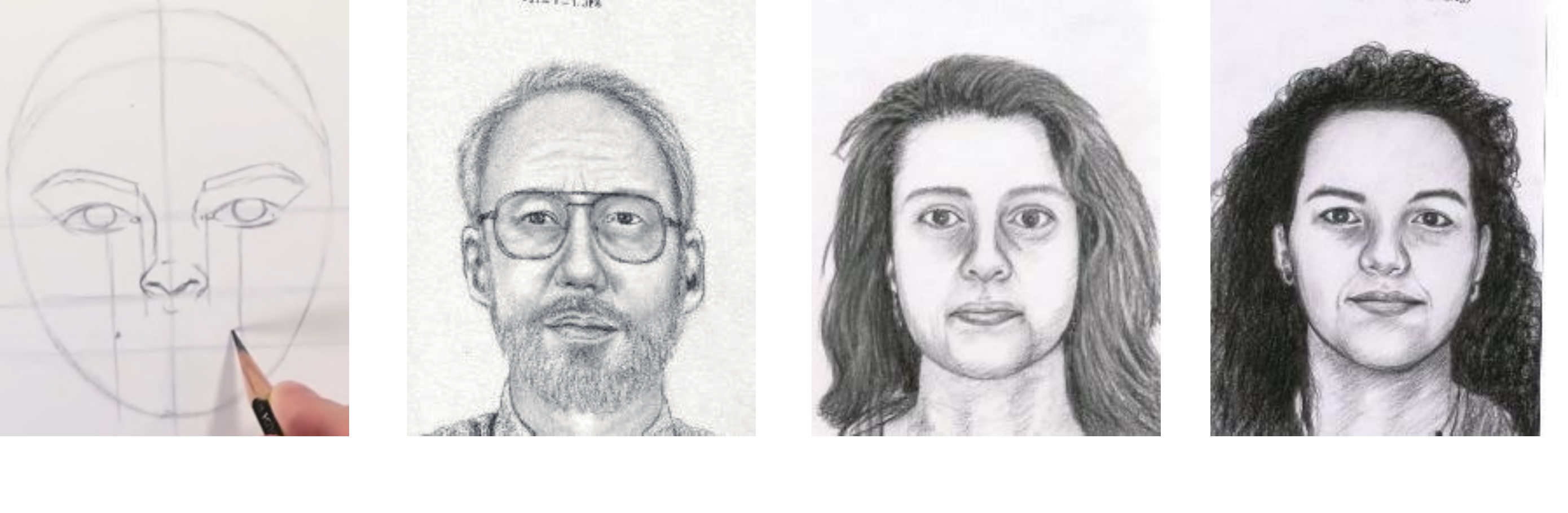}
    \put(3,0){(a) Outline}
    \put(30,0){(b) Light}
    \put(55,0){(c) Middle}
    \put(82,0){(d) Dark}
    \end{overpic}
    \caption{Creating various tones of the stroke by applying different pressure on the tip.
    (a) is the outline of the sketch. Strokes are added to this outline by perturbing
    the path of the tip to generate various sketches (b, c, d). From
    left to right these stroke tones are light, middle (in hair regions), dark.
    }\label{fig:PressureLevel}
\end{figure}

\subsection{Co-occurrence {\itshape Texture}}
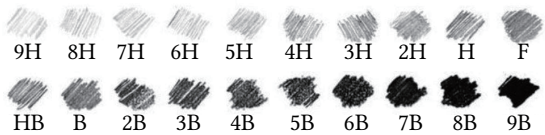
\begin{figure}[t]
\centering
    \begin{overpic}[width=.85\columnwidth]{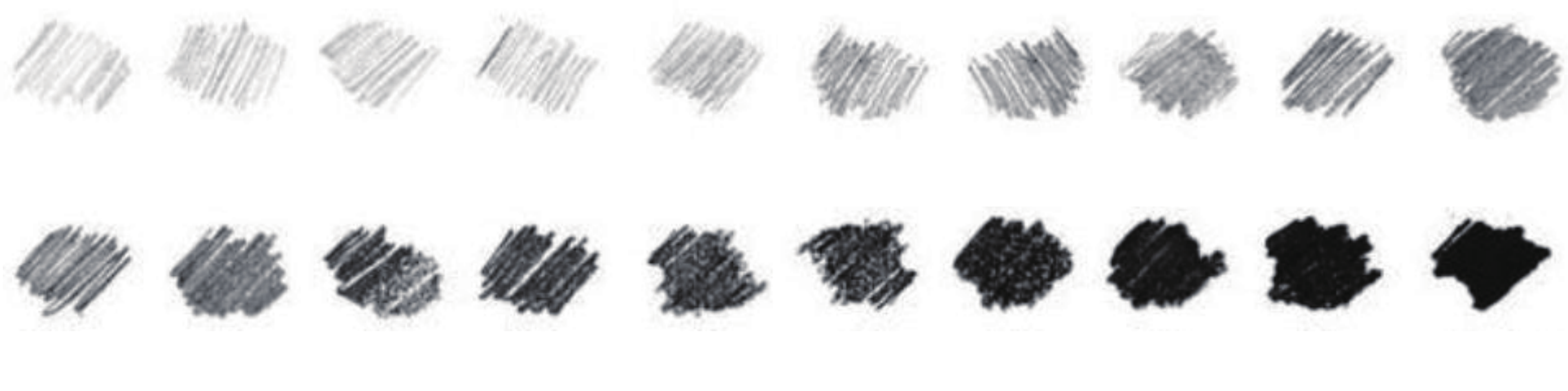}
    \put(2,14){9H}
    \put(12,14){8H}
    \put(21,14){7H}
    \put(31,14){6H}
    \put(41,14){5H}
    \put(52,14){4H}
    \put(63,14){3H}
    \put(73,14){2H}
    \put(84,14){H}

    \put(95,14){F}
    \put(2,1){HB}
    \put(13,1){B}
    \put(22,1){2B}
    \put(32,1){3B}
    \put(42,1){4B}
    \put(53,1){5B}
    \put(63,1){6B}
    \put(73,1){7B}
    \put(83,1){8B}
    \put(93,1){9B}
    \end{overpic}
    \caption{Pencil grades and their strokes.
    }\label{fig:PencilGrade}
\end{figure}

\subsubsection{Motivation.}
To uncover the secrets of face sketch and
explore quantifiable factors in style, we observed the basic
principles of the sketch and noted that the 
``graphite pencil grades''
and the ``pencil's strokes'' are the two fundamental elements in the sketch.

\emph{Graphite Pencil Grades.} In the European system, ``H'' \& ``B'' stand
for ``hard'' \& ``soft'' pencil, respectively. \figref{fig:PencilGrade}
illustrates the grade of graphite pencil from 9H-9B.
Sketch images are expressed through a limited medium (graphite pencil)
which provides no color. Illustrator Sylwia Bomba~\shortcite{Sylwia2015Sketching}
said that ``if you put your hand closer to the end of the pencil,
you have \emph{darker} markings. Gripping further up the
pencil will result in \emph{lighter} markings.''
In addition, after a long period of practice, artists will form their own \emph{fixed pressure} style.
In other words, the marking of the stroke can be varied by
varying the pressure on the tip (\figref{fig:PressureLevel} (c - d)).
It is important to note that different pressures on the tip will
result in various types of marking which is one of the quantifiable
factors called \textbf{gray tone}.

\begin{figure}[t!]
\centering
    \begin{overpic}[width=.6\columnwidth]{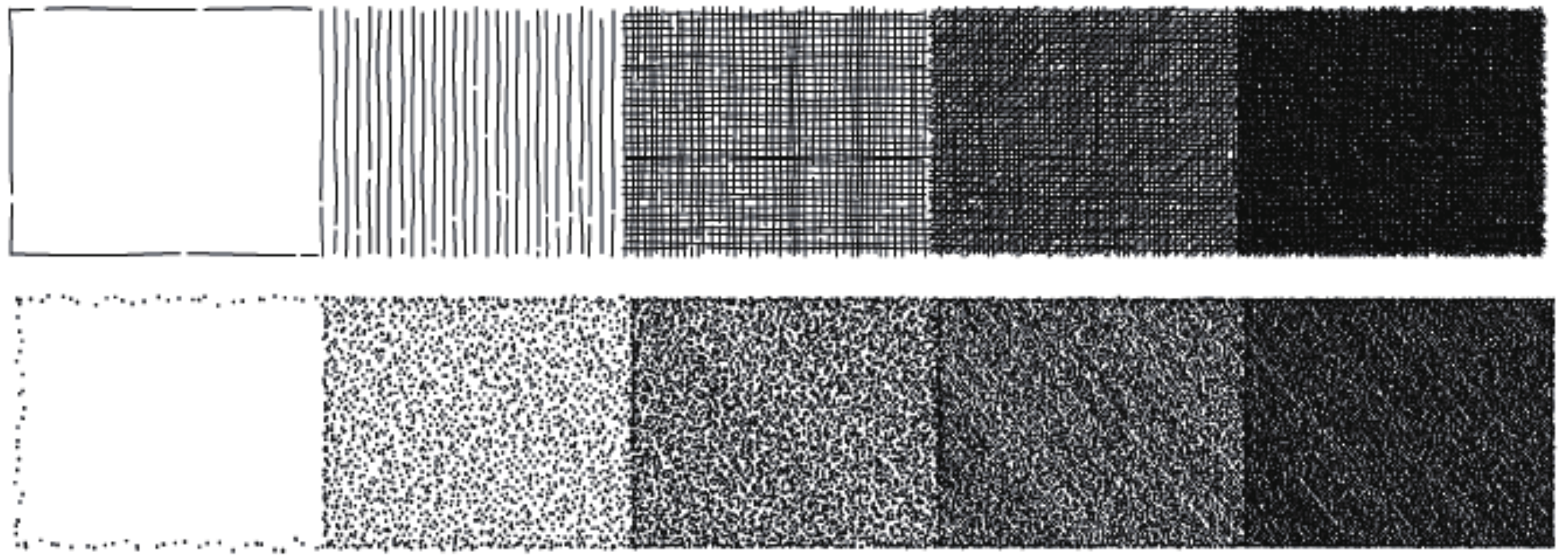}
    \end{overpic}
    \caption{Using stroke tones to indicate texture.
    The stroke textures used, from top to bottom, are: ``cross-hatching'',
    ``stippling''.
    The stroke attributes from left to right, are: sparse to dense.
    Images from ~\protect\cite{winkenbach1994computer}.}\label{fig:Texture}
\end{figure}

\emph{Pencil Strokes.}
Because all of the sketches are generated by moving a tip on the paper.
Hence, different paths of the tip along the paper will create various
stroke shapes. One example is shown in \figref{fig:Texture}; different spatial distributions
of the stroke have produced various textures (\eg \emph{sparse} or \emph{dense}).
Thus, the \textbf{stroke tone} is another quantifiable factor.

\subsubsection{Analysis}
In this section, we analyse the two quantifiable factors:
gray tone and stroke tone.

\textbf{Gray tone.} 
To reduce the effect of slight noise and 
over-sensitivity to imperceptible gray tone gradient changes in sketches, intensity quantization 
can be introduced during the evaluation of gray tone similarity.
Inspired by previous works~\cite{clausi2002analysis},
we can quantize the input sketch $I$ to $N_{l}$ different
grades to reduce the number of intensities  to be considered:
$I'  = \Omega(I)$.

\begin{figure}[t!]
\centering
    \begin{overpic}[width=.65\columnwidth]{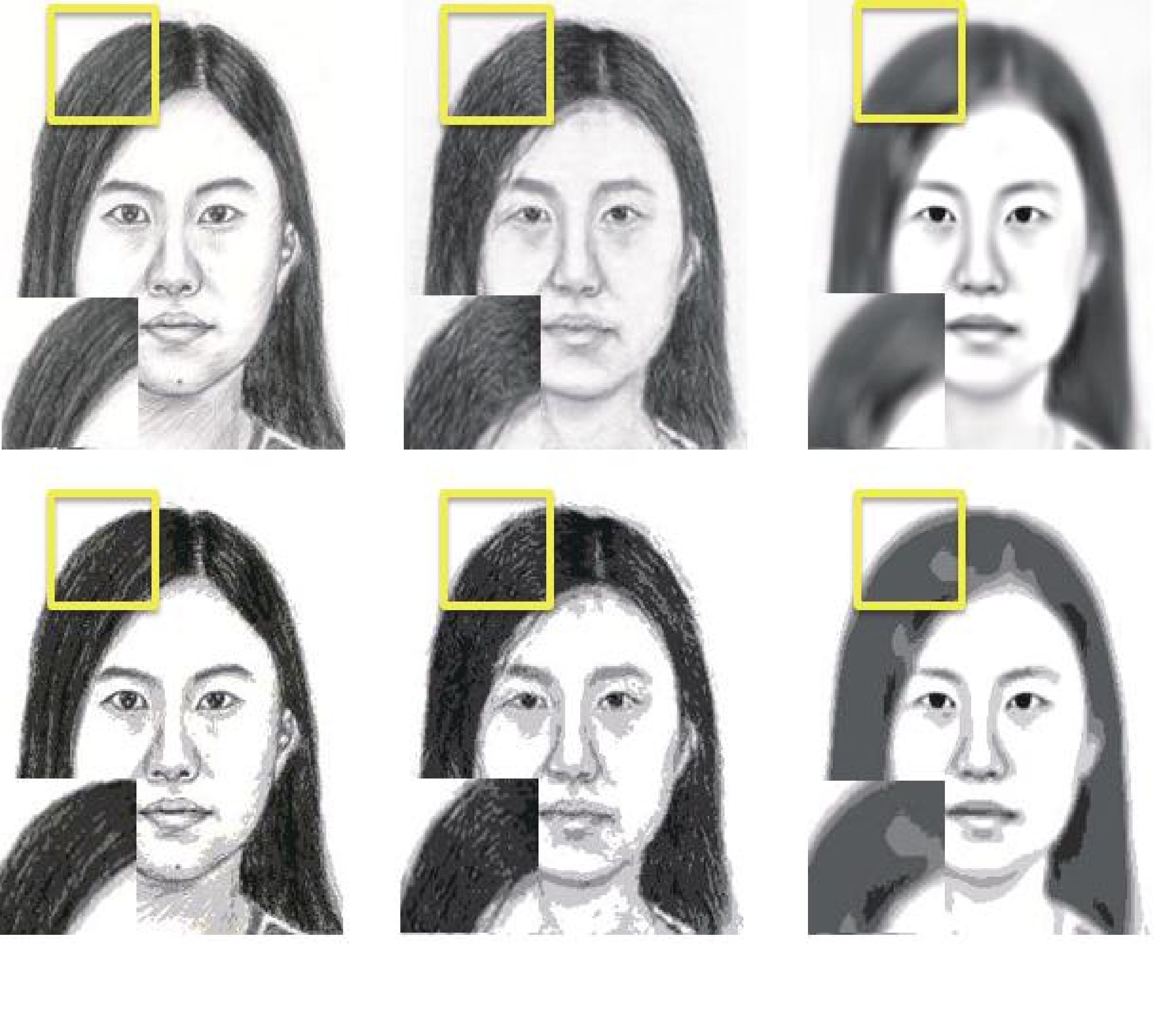}
    \put(3,0){(a) Sketch}
    \put(38,0){(b) GAN}
    \put(76,0){(c) LR}
    \put(-24,65){original}
    \put(-24,25){quantized}
    \end{overpic}
    \caption{Different stroke attributes result in
    different styles. The first row contains the original sketches.
    The second row contains the quantized sketches.
    The images of the first column are the GT sketches, and
    the images of the second and third columns are the synthesized
    sketches (GAN~\cite{isola2017image}, LR~\cite{wang2017data}). Texture of the hair region in (b)
    is more similar than that in (c) compared to the GT
    sketch in (a). Therefore, humans give (b) a higher score than (c).
    }\label{fig:Hair}
\end{figure}
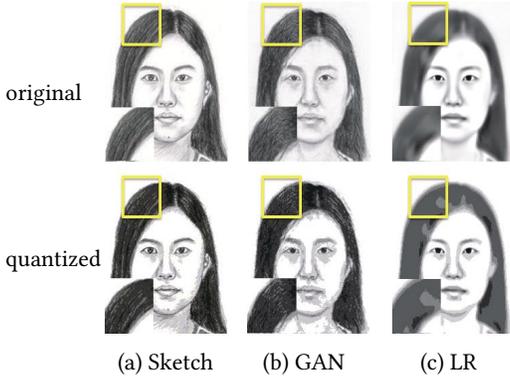

A typical example of such quantization is shown in \figref{fig:Hair}.
Human beings will consistently rank (b) higher than (c) before and after
quantizing the input sketches when evaluating the style similarity.
Although, the quantization technology may introduces artifacts.
However, our experiments (\secref{sec:discussion} of the quantized Scoot measure)
show that this process can
reduce sensitivity to minor intensity variations, thus achieving the best overall
performance, and also reducing the computational complexity.

\textbf{Stroke tone.}
Although we can efficiently capture the gray tone by
quantizing the input sketch to $N_{l}$ grades
we notice that the stroke tone (\eg dense \& sparse) is
also important to formulate the sketch style.
Stroke tone and gray tone are not independent concepts; rather,
they bear an inextricable relationship to one another.
Both gray tone and stroke tone are innate properties of the sketches.
The gray tone is based on the varying strokes of gray-scale
in a sketch image, while the stroke tone is concerned with
the spatial (statistical) distribution of gray tones.

This is demonstrated in \figref{fig:Hair},
which presents the ground-truth sketch (a) and two other
synthesis results (b) \& (c). These two images are
the results of SOTA face synthesis algorithms.
Intuitively, result (b) is better than (c), since (b)
preserves the texture of the hair and details in
the face while (c) presents a smooth result and has lost much of the
sketch style.
In other words, (b) and (c) have different stroke patterns.
The co-occurrence stroke pattern of (b) has a higher similarity than the sketch in (c).

\subsubsection{Implementation.}\label{sec:implementation}
After confirming the two quantifiable factors,
we start to describe our implementation details.
To simultaneously extract statistics about the ``stroke tone''
and their relationship to the surrounding ``gray tone'', we need
to characterize their \emph{spatial interrelationships}.
Theoretically, the sketches are quite close to textures.
As a consequence, our goal is to capture the \emph{\textbf{co-occurrence texture}}
in the sketch. Such co-occurrence properties help to capture the patterns of the face sketch (especially in the hair region). 
Similar to the general technology~\cite{haralick1973textural}, 
we built the co-occurrence gray tones in a matrix 
at  distance 
$d=(\Delta x, \Delta y)$
to encode the dependency of the gray tones.
Specifically, this matrix $M$ is defined as:
\begin{equation}\label{equ:HaralickMatrix}
M_{(i,j)|d}=\sum_{y=1}^{H}\sum_{x=1}^{W}
\begin{cases}
1,\mbox{ if } I'_{x,y}=i \ \mbox{and} \ I'_{(x,y)+d}=j \\
0,\mbox{ otherwise}
\end{cases}
,
\end{equation}
where $i$ and $j$ denote the gray value; $x$ and $y$ are the spatial positions
in the given quantized sketch $I'$; $I'_{x,y}$ denotes the gray value of $I'$ at position $(x,y)$;
$W$ and $H$ are the width and height of the sketch $I'$, respectively.

To extract the style features we tested the three most widely-used~\cite{haralick1979statistical} 
statistics: Homogeneity, Contrast, and Energy.

\textbf{Homogeneity} reflects how much the texture changes in local regions,
it will be high if the gray tone of each pixel pair is similar.
The homogeneity is  defined  as:
\begin{equation}\label{equ:Homogeneity}
    h = \sum_{j=1}^{N_l}\sum_{i=1}^{N_l}{\frac{M_{(i,j)|d}}{1+|i-j|}},
\end{equation}

\textbf{Contrast} represents the difference between a pixel in $I'$ and its neighbor summed over
the whole sketch. This reflects that a low-contrast sketch is not characterized by
low gray tones but rather by low spatial frequencies. The contrast is highly correlated with
spatial frequencies. The contrast equals 0 for a constant tone sketch.
\begin{equation}\label{equ:Contrast}
    c = \sum_{j=1}^{N_l}\sum_{i=1}^{N_l}{|i-j|^2M_{(i,j)|d}}
\end{equation}

\textbf{Energy} measures textural uniformity. When only similar gray tones of pixels occur in
an sketch ($I'$) patch, a few elements in $M$ will be close to 1, while others will be
close to 0.  Energy will reach the maximum if there is only one gray tone in a sketch ($I'$) patch. Thus, high energy corresponds to
the sketch's gray tone distribution having either a periodic or constant form.
\begin{equation}\label{equ:Uniformity}
    e = \sum_{j=1}^{N_l}\sum_{i=1}^{N_l}{(M_{(i,j)|d})^2}
\end{equation}

\textbf{CE (Contrast and Energy) Feature.}
Our experiments (see \tabref{table:FeatureSelection}) show that the combination of two statistics
Contrast and Energy achieves the best average performance.
In this work, we concatenate the two statistics as the CE feature to represent the
style of the input sketch $I'_{s}$.

\subsection{Spatial {\itshape Structure}}
Note that the matrix $M$ \eqnref{equ:HaralickMatrix} we built is based at the ``image-level''.
However, the weakness of this approach is that it only captures the global statistics, and
the structure of the sketch is ignored. Therefore, for instance, it would be
difficult to distinguish men and women with the same hairstyle.

To holistically represent the spatial structure, we follow the spatial 
envelope method~\cite{oliva2001modeling} to capture the statistics from the ``block-level''
\emph{\textbf{spatial structure}} in the sketch.
Firstly, we divide the whole sketch image into a $k \times k$ grid of $k^2$ blocks\footnote{
We set $k = 4$ to achieve the best performance in our all experiments.
} before extracting the CE feature.
Our experiments demonstrate that the process can help to derive content information.
Second, we compute the co-occurrence matrix $M$ for all blocks and
normalize each matrix such that the sum of its components is 1.
Finally, we concatenate the statistics of all the $k^2$ blocks into a $2k^2$ dimension vector
\begin{math}
  \overrightarrow{\Phi}(I'_{s}|d).
\end{math}

Note that each of the statistics is only based on a single direction (\eg $90^o$, that is d = (0, 1)), since the direction
of the spatial distribution is also very
important to capture the style such as ``hair direction'',
``the direction of shadowing strokes''. To exploit this observation
for efficiently extracting the \emph{stroke direction} style, we compute
the average feature $\overrightarrow{\Psi}(I'_{s})$ of four orientations\footnote{
Due to the symmetry of the co-occurrence matrix $M(i,j)$,
the statistical features in 4 orientations are actually equivalent to 8 neighbors direction
at 1 distance. Empirically, we set 4 orientations
$d_{i}\in \{(0,1), (-1,1), (-1,0), (-1,-1)\}$
to achieve the best performance.
} vectors to capture more directional information:
\begin{equation}\label{equ:Homogeneity}
    \overrightarrow{\Psi}(I'_{s})= \frac{1}{4}\sum_{i=1}^{4}{\overrightarrow{\Phi}}(I'_{s}|d_{i}),
\end{equation}
where $d_{i}$ denotes the $i$th direction and the average CE feature $\overrightarrow{\Psi}(I'_{s})\in \mathbb{R}^{2k^2}$.
\subsection{Scoot measure}
Having obtained the style feature vectors of the GT sketch $Y$ and synthesis sketch $X$, 
we use the Euclidean distance between their feature vectors to evaluate their style similarity.
Thus, our style \textbf{Scoot-measure}\footnote{
In all of our experiment, we set $N_{l} = 6$ grades
to achieve the best performance.} can be defined as:

\begin{equation}\label{equ:Scoot}
    E_{s} = \frac{1}{1+\left\|\overrightarrow{\Psi}(X'_{s})-\overrightarrow{\Psi}(Y'_{s})\right\|_2},
\end{equation}
where $\left\|\cdot\right\|_2$ denotes $l_2$-norms. 
$X'_{s}, Y'_{s}$ denote the quantized $X_{s}, Y_{s}$, respectively. 
$E_s=1$ corresponds to identical style.
Using this measure to evaluate the three synthesized face sketches in \figref{fig:FirstExample}, we can correctly rank the sketches consistent with 
the human ranking.

\section{Experiment}\label{sec:experiment}
In this section, we compare our Scoot-measure with
8 measures including 4 popular measures (FSIM, SSIM, VIF,
GMSD) and 4 alternative baseline measures
(GLRLM~\cite{galloway1974texture}, Gabor~\cite{gabor1946theory},
Canny~\cite{canny1987computational}, Sobel~\cite{sobel1990isotropic}) which related with
texture or edge. For the four baseline measures, we
only replace our CE feature (describe in \secref{sec:implementation}) with these features.

\textbf{Meta-measure.}
To test the quality of our measure, we adopt the {\em meta-measure} methodology.
The meta-measures consist of assuming some plausible hypotheses about the
results and assessing how well each measure reflects these hypotheses~\cite{pont2013measures,fan2017structure}.
We design 4 novel meta-measures based on our plausible hypotheses
(\secref{sec:Introduction}) and the experiment results are summarized in
\tabref{table:MetaMeasureResults}.

\textbf{Dataset.}
All of the meta-measures were performed on 2 widely used databases:
\textbf{CUFS}~\cite{wang2009face}
and \textbf{CUFSF}~\cite{zhang2011coupled}. The CUFS database includes
606 face photo-sketch pairs consisting of 3 sub-sections:
CUHK Student (188 pairs), XM2VTS (295 pairs) and Purdue AR
(123 pairs). The CUFSF database contains 1194 pairs of
persons from the FERET database~\cite{phillips2000feret}.
There is one face photo and one GT sketch
drawn by the artist for each person in both CUFS and CUFSF databases.
Following~\cite{wang2018random}, we use 338 pairs (CUFS) and 944 pairs
(CUFSF) test set images to conduct our experiments.

\begin{figure}[t!]
\centering
    \begin{overpic}[width=\columnwidth]{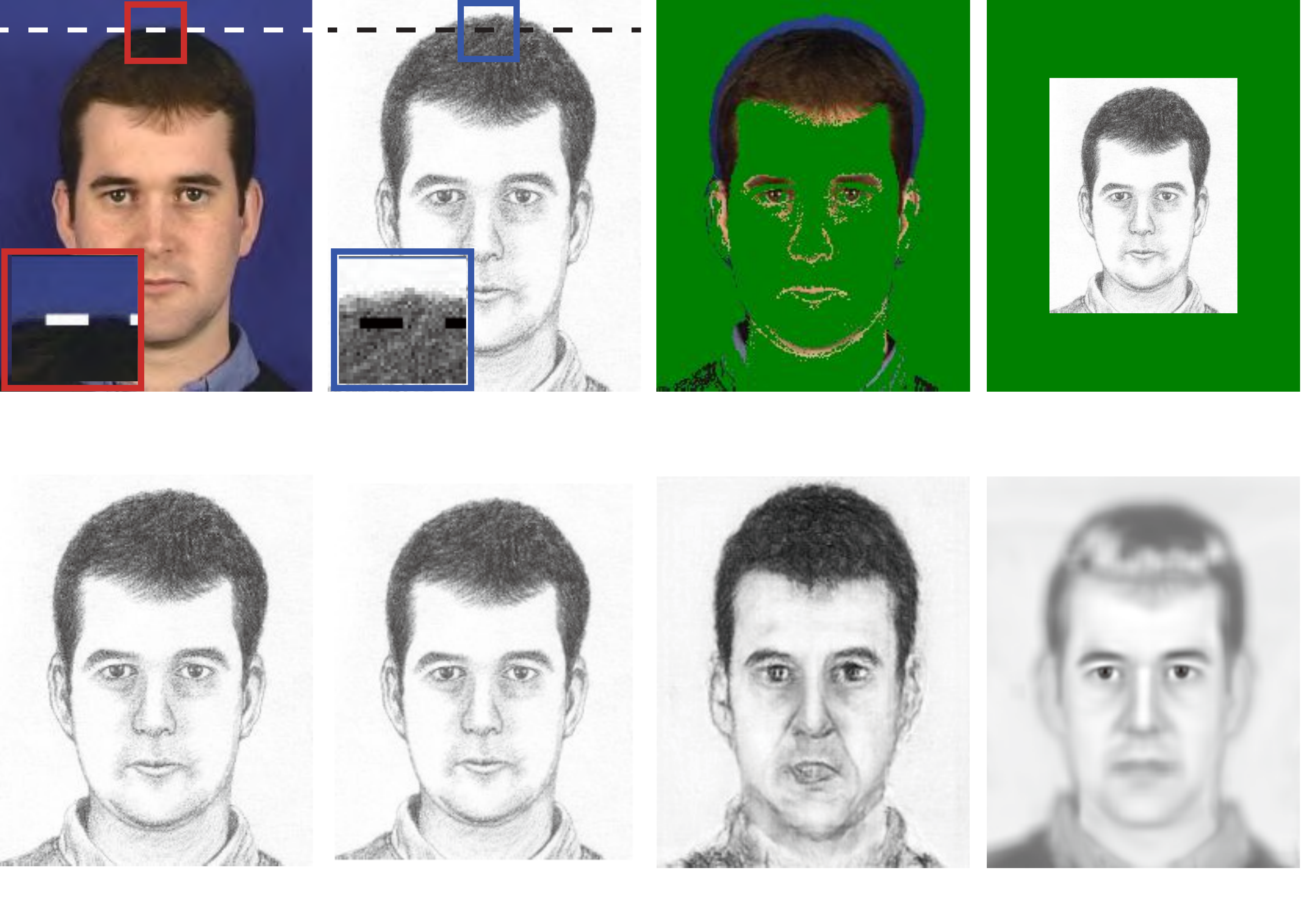}
    \put(5,36){(a) Photo}
    \put(32,36){(b) GT} 
	\put(51,36){(c) Differenced}
    \put(77,36){(d) downsized}

    \put(6,0){(e) GT}
    \put(28,0){(f) Resized}
    \put(56,0){(g) GAN}
    \put(83,0){(h) LR}
    \end{overpic}
    \caption{Meta-measure 1. 
    (a) is the original photo. (b) is the GT sketch drawn by the artist. 
    (c) is the boundary difference between (a) \& (b). (d) is the 
    downsized image from (b).
    The ranking of an evaluation measure should be insensitive
    to slight resizing of the GT. While GT (e) \& 5 pixels scaled-down GT (f)
    differ slightly, widely-used measures (\eg VIF, SSIM, and GMSD) switched the ranking order of the two
    synthesized sketches (g) \& (h) when using the different versions of GT referenced 
    (e, f).
    In contrast, our Scoot measure consistently ranked (g) higher than (h).
    }\label{fig:MM1}
\end{figure}

\begin{figure}[t]
\centering
    \begin{overpic}[width=.95\columnwidth]{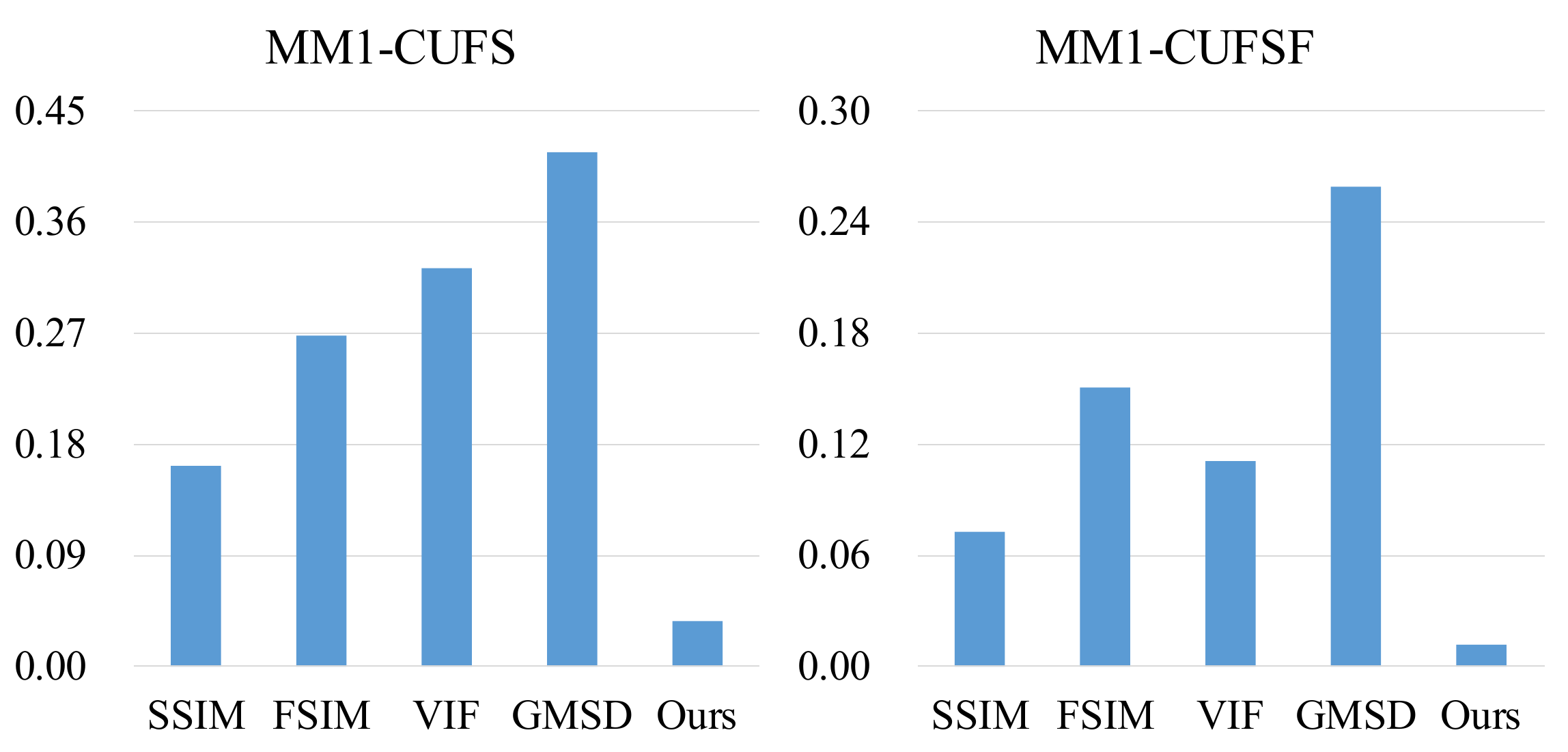}
    \end{overpic}
    \caption{Meta-measure 1 results:
    The lower the result is, the more stable an evaluation measure
    is to slight resizing.}\label{fig:MM1Result}
\end{figure}

\textbf{Sketch Synthesis Results.}
Sketch synthesis results were generated for each test image
using 10 SOTA models (FCN~\cite{zhang2015end},
GAN~\cite{isola2017image}, LLE~\cite{liu2005nonlinear},
LR~\cite{wang2017data}, MRF~\cite{wang2009face},
MWF~\cite{zhou2012markov}, RSLCR~\cite{wang2018random},
SSD~\cite{song2014real}, DGFL~\cite{zhu2017deep}, BFSS~\cite{wang2017bayesian}).

\subsection{Meta-measure 1: Stability to Slight Resizing}\label{sec:Insensitive}
The first meta-measure specifies that the rankings of
synthetic results should not change much with
slight changes in the GT sketch. Therefore, we perform a slight
5 pixels downsizing of the GT by using nearest-neighbor interpolation.

\figref{fig:MM1} gives an example. The hair of the GT
sketch in (b) drawn by the artist has a slight size discrepancy
compared to the photo (a). We observe that about 5 pixels
deviation (\figref{fig:MM1} c) in the boundary is common.
While the two sketches (e) \& (f) are almost identical, commonly-used
measures including SSIM, VIF, and GMSD switched the ranking of the two 
synthesized results (g, h) when using (e) or (f).
However, our measure consistently ranked (g) higher than (h).

Here, we applied the $\theta=1-\rho$~\cite{best1975algorithm} measure
to test the measure ranking stability before and after the
GT downsizing was performed. \figref{fig:MM1Result} and \tabref{table:MetaMeasureResults} show the results:
the lower the result is, the more stable an evaluation measure is to slightly
downsizing. We can see a significant ($\approx$ 88\% and 92\%) improvement
over the existing SSIM, FSIM, GMSD, and VIF measures in both the CUFS and CUFSF databases.
These improvements are mainly because our evaluation measure considers ``block-level''
features rather than ``pixel-level''.

\subsection{Meta-measure 2: Rotation Sensitivity}\label{sec:rotation}

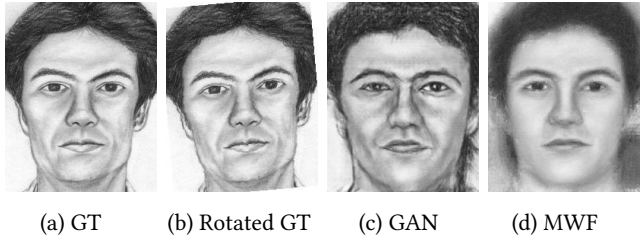
\begin{figure}[t!]
\centering	
    \begin{overpic}[width=\columnwidth]{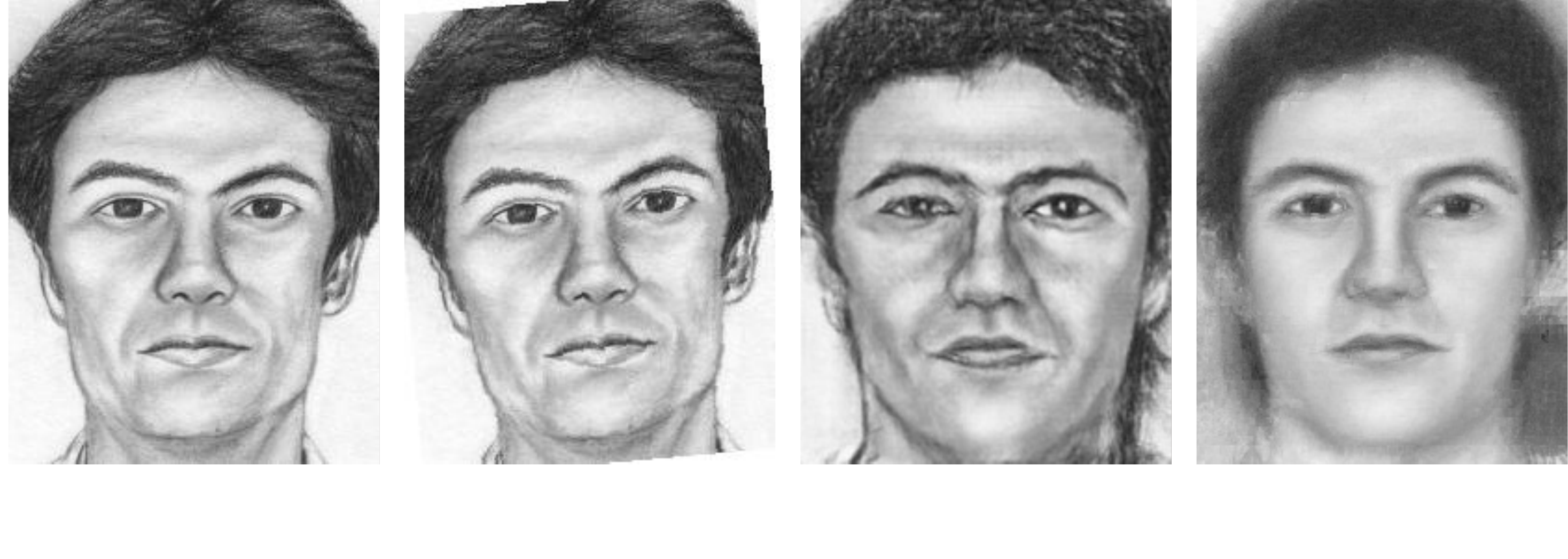}

    \put(6,0){(a) GT}
    \put(26,0){(b) Rotated GT}
    \put(56,0){(c) GAN}
    \put(80,0){(d) MWF}
    \end{overpic}
    \caption{Meta-measure 2.
    The ranking of an evaluation measure should be insensitive
    to slight rotation of the GT.
    While GT (a) and $5^o$ rotated GT (b)
    has the same style, all of the current measures switched
    the ranking order of the two synthesized sketches (c) \& (d),
    relying on the different GT referenced.
    Oppositely, our Scoot measure consistently ranked (c) higher than (d).
    }\label{fig:MM2}
\end{figure}

In real-wold situations, sketches drawn by artists may also have slight rotations compared to
the original photographs. Thus, our second meta-measure verifies the sensitivity of GT rotation
for the evaluation measure.
We did a slight counter-clockwise rotation ($5^o$) for each GT sketch.
\figref{fig:MM2} shows an example.
When the GT (a) is switched to the slightly rotated GT (b), the ranking results should not change much.

\begin{figure}[t]
\centering
    \begin{overpic}[width=.95\columnwidth]{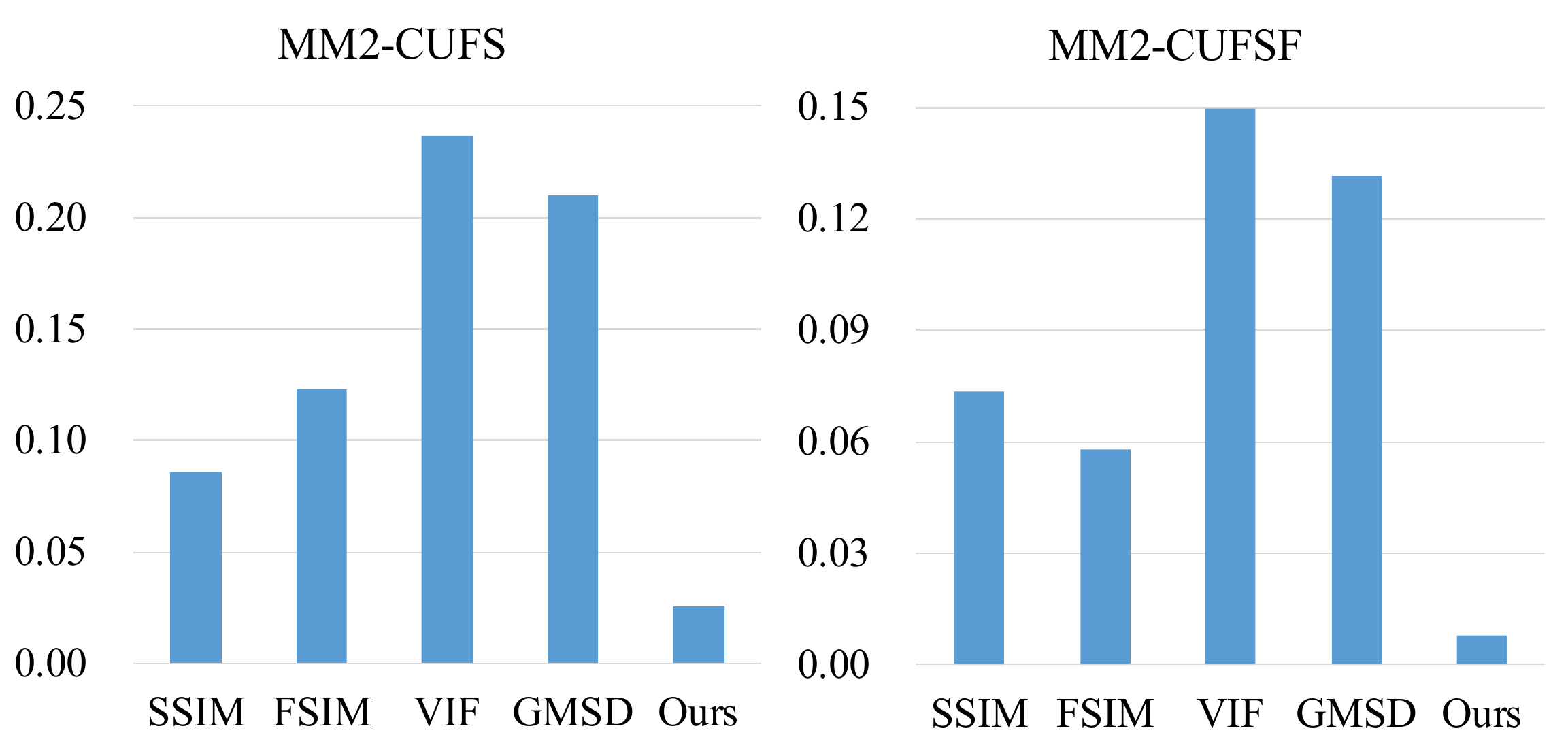}
    \end{overpic}
    \caption{Meta-measure 2 results:
    The lower the result is, the more stable an evaluation measure
    is to slight rotation.}\label{fig:MM2Result}
\end{figure}

We got the ranking results for each measure by using GT sketches and slightly rotated GT sketches separately
and applied the same measure ($\theta$) as meta-measure 1 mentioned to evaluate the rotation sensitivity.
The sensitivity results are shown in \figref{fig:MM2Result}, \tabref{table:MetaMeasureResults}. Thanks to
our use of ``block-level'' statistics, our measure again significantly outperforms the current measures over the CUFS and CUFSF databases. 

\subsection{Meta-measure 3: Content Capture Capability}\label{sec:Content}

\begin{figure}[t!]
\centering
    \begin{overpic}[width=.75\columnwidth]{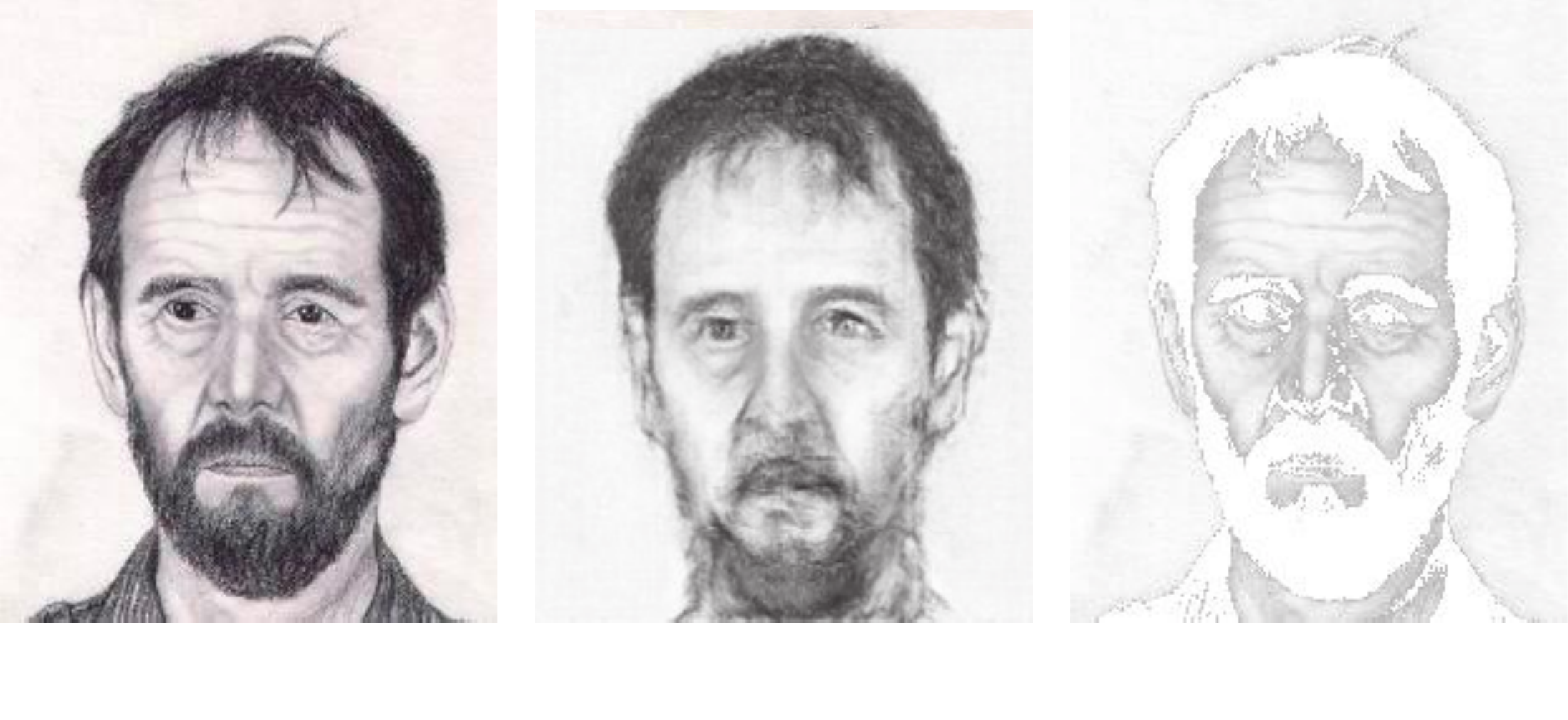}
    \put(10,0){(a) GT}
    \put(37,0){(b) Synthesis}
    \put(75,0){(c) Light}
    \end{overpic}
    \caption{Meta-measure 3.
    We use a simple threshold to separate the GT sketch (a)
    into dark strokes and light strokes (c). From the
    style similarity perspective, a good evaluation measure should
    prefer the SOTA synthesized sketch (b) which contains the main texture and
    structure over the result only contains light strokes and has lost the main facial features (c).
    }\label{fig:MM3}
\end{figure}

The third meta-measure describes that a good evaluation measure
should not only evaluate the style similarity but also can
capture the content similarity.
\figref{fig:MM3} presents an example.
We expect that an evaluation measure should prefer the SOTA
synthesized result over the \textbf{light strokes}\footnote{
To test the third meta-measure, we use a simple threshold of gray scale
(\eg 170) to separate the sketch (\figref{fig:MM3} GT) into darker strokes
\& lighter strokes. The lighter strokes image loses the main
texture features of the face (\eg hair, eye, beard), resulting in an incomplete sketch.
} results. To our surprise, only our measure gives the correct order.

\begin{figure}[t!]
\centering
    \begin{overpic}[width=.95\columnwidth]{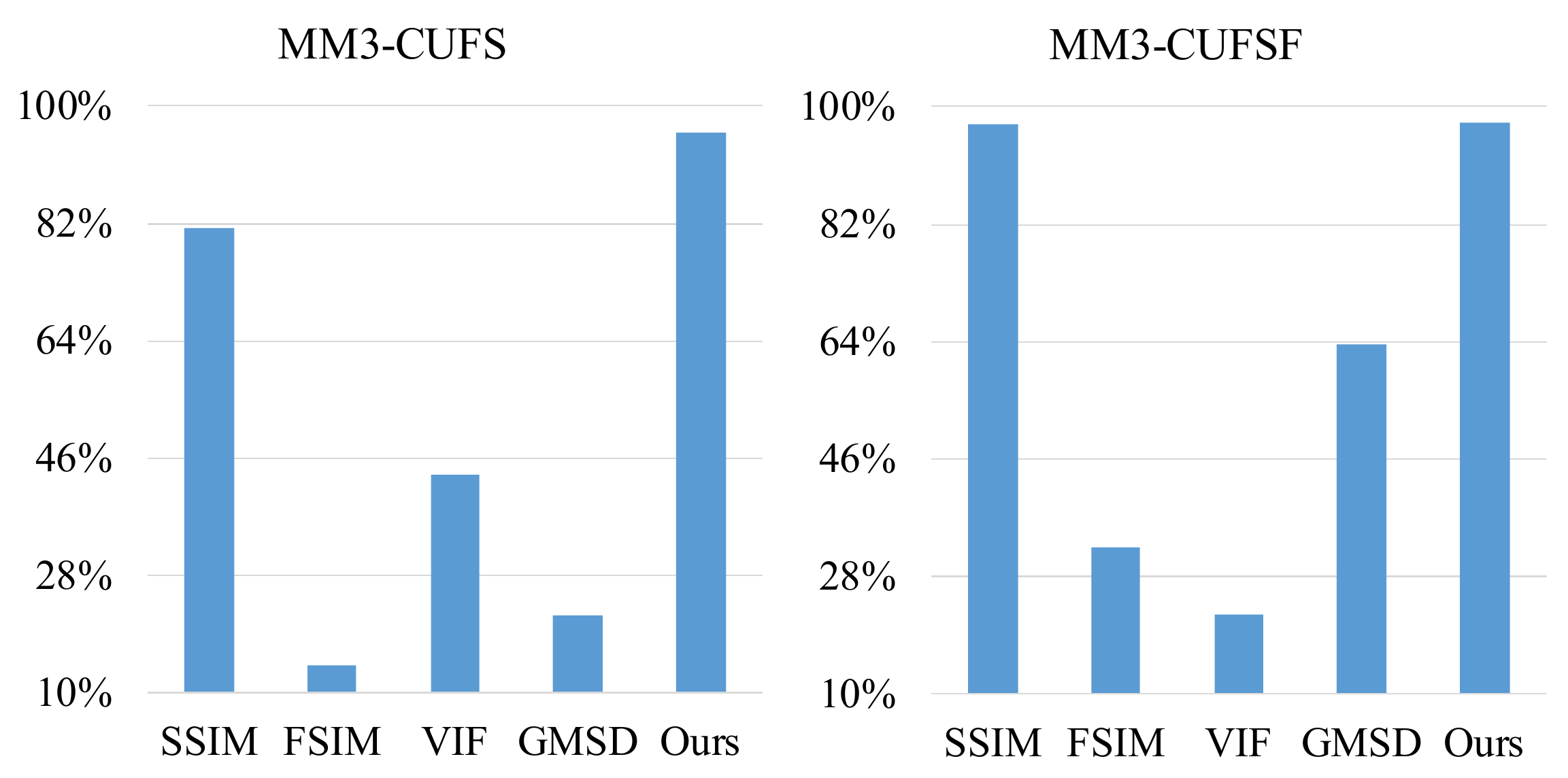}
    \end{overpic}
    \caption{Meta-measure 3 results:
    The higher the result, the stronger is the  holistic content capture
    capability. Our measure achieves the best
    performance in both the CUFS and CUFSF databases.}\label{fig:MM3Result}
\end{figure}

We use the \emph{mean score} of 10 SOTA synthesis
algorithms (FCN, GAN, LLE, LR, MRF, MWF, RSLCR, SSD, DGFL, BFSS).
The mean score is robust to situations in which a certain model
generates a poor result. We recorded the number of times
the mean score of SOTA synthesized algorithms is higher than a light stroke's score.
The results are shown in \figref{fig:MM3Result} \& \tabref{table:MetaMeasureResults}.
In the CUFS database we can see a great improvement over the
most widely-used measures. A slight improvement is also achieved
for the CUFSF database.

The reason why other measures fail in this meta-measure is shown in \figref{fig:MM3}.
In terms of pixel-level matching, it is obvious that the regions where dark strokes are
removed are totally different from the corresponding parts in (a).
But at other positions the pixels are identical to the GT.
 Previous measures only consider ``pixel-level'' matching,  and will rank the light strokes sketch higher.
However, the synthesized sketch (b) is better than the light one (c) in terms of both style and content.

\subsection{Meta-measure 4: Human Judgment}\label{sec:Judgment}
The fourth meta-measure specifies that the ranking result
according to an evaluation measure should agree with the human ranking.
As far as we know, there is no such human ranking sketch database publicly
available.

\textbf{Source images.} As mentioned
in \secref{sec:experiment}, our source sketch synthesized
results are from the two widely-used datasets: CUFS and CUFSF and
10 SOTA sketch synthesis algorithms. Thus, we have
3380 (10 $\times$ 338) and 9440 (10 $\times$ 944) synthesized results
for CUFS and CUFSF, respectively.

\textbf{Human judgment.} For each photo of the CUFS and CUFSF databases,
we have 10 sketches synthesized by 10 SOTA algorithms. Due to the
similarity of the synthesized results of some algorithms, it is difficult for
viewers to judge the ranking. Following~\cite{wang2016evaluation}, we trained 
the viewers to assess synthesized sketch quality based on two criterion:
texture similarity and content similarity. To minimize the ambiguity of human ranking,
we asked 5 viewers to select 2 synthesized sketches for each photo
through the following 3 stage processes:

\vspace{8pt}
\textbf{i)} We let the first group of viewers select 4 out of 10 sketches for each photo.
The 4 sketches should consist of two good and two bad ones.
Thus we are left with 1352 (4 $\times$ 338) and 3776 (4 $\times$ 944) sketches for CUFS and CUFSF, respectively.

\vspace{8pt}
\textbf{ii)} For the 4 sketches in each photo, the second group of viewers is further asked to select 3 sketches for which they can rank them easily.
Based on the voting results of viewers, we pick out the 3 most selected sketches.

\begin{figure}[t!]
\centering
    \begin{overpic}[width=\columnwidth]{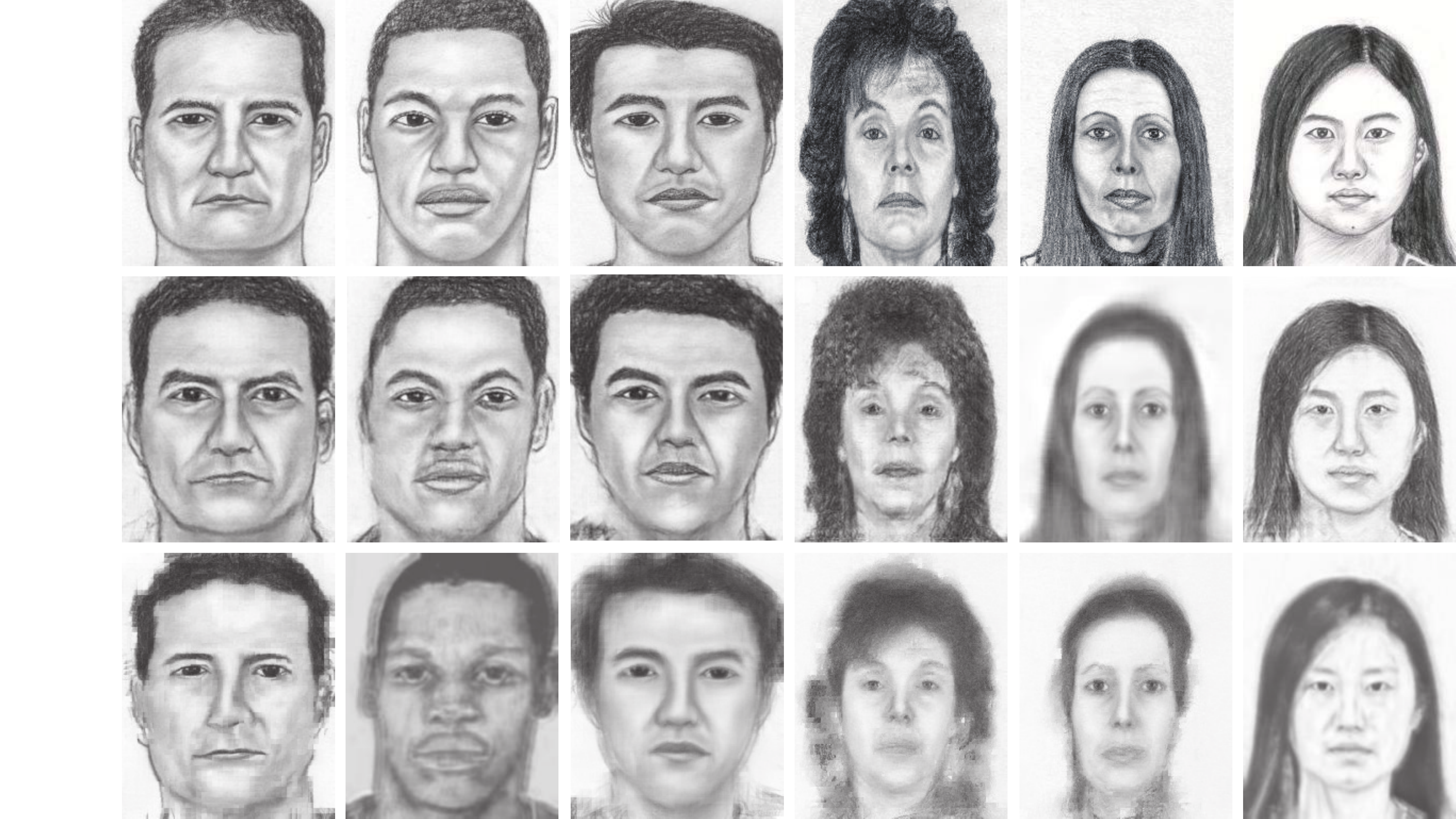}
    \put(1,45){GT}
    \put(1,25){$1^{st}$}
    \put(1,8){$2^{nd}$}
    \end{overpic}
    \caption{Meta-measure 4.
    Sample images from our human ranked database.
    The first row is the GT sketch, followed by
    the first and second ranked synthesis result. 
    Please refer the supplementary material for more images.
    }\label{fig:MM4}
\end{figure}

\vspace{8pt}
\textbf{iii)} The last group of viewers are asked to pick out a pair of sketches that are most obvious to rank.
Note that we have 5 volunteers involved in the whole process for cross-checking the	 ranking.
Finally, we create two new human ranked datasets:
\textbf{CUFS-Ranked} and \textbf{CUFSF-Ranked} (see the \emph{supplementary} for all images in these two datasets)~\footnote{They 
include 776 \& 1888 human ranked images, respectively. 
The two datasets will be made publicly available.}.
\figref{fig:MM4} shows the human ranked examples.

\begin{figure}[t!]
\centering
    \begin{overpic}[width=.95\columnwidth]{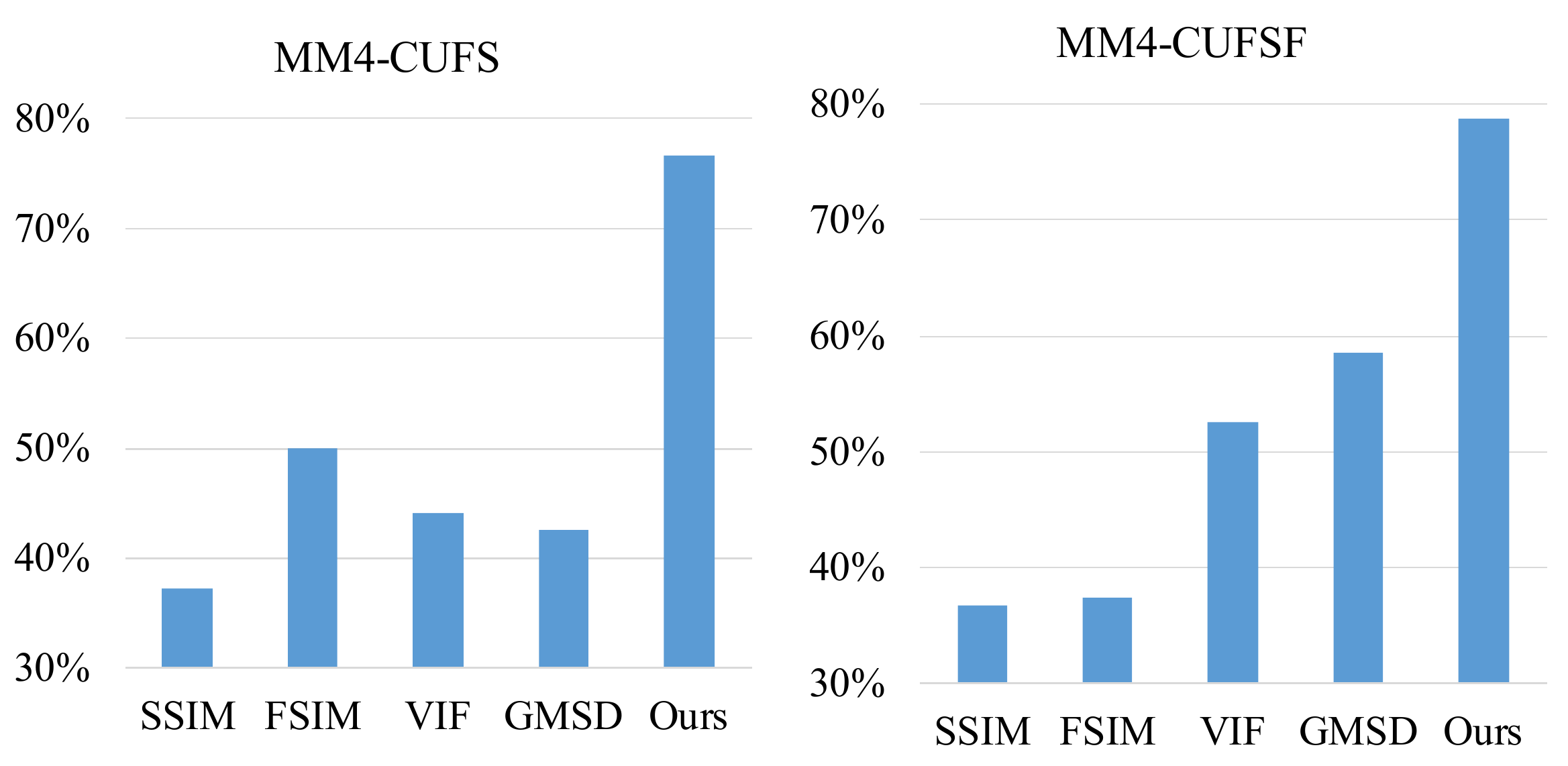}
    \end{overpic}
    \caption{Meta-measure 4 results:
    The higher the result, the more consistency an evaluation measure is 
    to human judgment. Our measure achieves the best
    performance in both CUFS and CUFSF databases.}\label{fig:MM4Result}
\end{figure}

Here, we examined the 
the consistency between
human ranking and evaluation measure ranking.
Results are given in \figref{fig:MM4Result},
which shows a significant (about 26.3\%) improvement over
the best prior measure in CUFS. This improvement is due to our consideration of style
similarity which human perception considers as an essential factor when evaluating sketches.

\section{Discussion}\label{sec:discussion}
In this section, we will discuss the combination of statistics in
\secref{sec:featureAnalysis} and other alternative features in \secref{sec:otherFeature}. These discussions indicate that 
the pair-wise ``co-occurrence'' properties are quite important for the style similarity evaluation for FSS.

\begin{table}[t!]
  \centering
  \footnotesize
  \renewcommand{\tabcolsep}{1.5mm}
  \caption{Quantitative comparison of different combination of 3 statistical features on our 4 meta-measures.
  MM: meta-measure. The best balanced performance is highlighted in {\textbf{\color{black}{bold}.}}
  These differences are all statistically significant at the $\alpha< 0.05$ level.}\label{table:FeatureSelection}
  \begin{tabular}{l|cccc|cccc}
  \toprule
  &\multicolumn{4}{c|}{CUFS} & \multicolumn{4}{c}{CUFSF}\\
  \cmidrule{2-9}
  Measure & MM1 & MM2 & MM3 & MM4 & MM1 & MM2 & MM3 & MM4\\
  \toprule
  $h$   & 0.007 & 0.005 & 61.5\% & 77.5\% & 0.003 & 0.003 & 79.1\% & 77.8\%\\
  $e$   & 0.200 & 0.104 & 98.5\% & 73.1\% & 0.044 & 0.026 & 99.2\% & 77.4\%\\
  $c$   & 0.010 & 0.007 & 54.4\% & 74.6\% & 0.009 & 0.006 & 64.7\% & 73.4\%\\
  $c + h$ & 0.011 & 0.007 & 60.1\% & 74.6\% & 0.007 & 0.005 & 78.1\% & 73.7\%\\
  $\textbf{c}$ $+$ $\textbf{e}$& \textbf{0.037} & \textbf{0.025} & \textbf{95.9\%} & \textbf{76.3\%}
  & \textbf{0.012} & \textbf{0.008} & \textbf{97.5\%} & \textbf{78.8\%}\\
  $h + e$ & 0.156 & 0.088 & 97.9\% & 75.7\% & 0.030 & 0.017 & 98.8\% & 80.3\%\\
  $h + e + c$&0.034 & 0.024 & 95.9\% & 76.3\% & 0.011 & 0.008 & 97.4\% & 78.7\%\\
  \bottomrule
  \end{tabular}
\end{table}

\subsection{Statistical Feature Selection}\label{sec:featureAnalysis}
According to \secref{sec:methodology}, we considered three widely-used features of co-occurrence matrices: homogeneity ($h$), contrast ($c$) and energy ($e$).
In order to achieve the best performance, we need to explore the
best combination of these statistical features.
We have applied our four meta-measures to test the performance of our measure using each single feature, each feature pair and the combination of all three features.

The results are shown in \tabref{table:FeatureSelection}. 
All possibilities ($h, e, c, c + e, h +e , c + h, h + e + c$) 
perform well in MM4 (human judgment). 
$h$ and $c$ are insensitive to resizing (MM1) and rotation (MM2), while they are not good at content capture (MM3). $e$ is the opposite compared to $h$ and $c$. Thus, using a
single feature is not good.
The results of combining two features show that if $h$ is combined with $e$, the sensitivity to resizing and rotating will still be high, while partially overcoming the weakness of $e$. The performance of $h + e + c$ shows no improvement compared to the combination of $c + e$. Previous work in~\cite{baraldi1995investigation} also found energy and contrast to be the most efficient features for discriminating textural patterns.
Therefore, we choose contrast and energy as our final combination ``CE'' feature to extract the style features.

\begin{table*}[t]
  \centering
  \caption{Quantitative comparison of different evaluation measures on our 4 meta-measures.
  MM: meta-measure. The top measure for each MM are shown in boldface.
  These differences are all statistically significant at the $\alpha< 0.05$ level.}\label{table:MetaMeasureResults}
  \begin{tabular}{l|cccc|cccc}
  \toprule
  &\multicolumn{4}{c|}{CUFS (3380 synthesized images)} & \multicolumn{4}{c}{CUFSF (9440 synthesized images)}\\
  \cmidrule{2-9}
  Measure & MM1 & MM2 & MM3 & MM4 & MM1 & MM2 & MM3 & MM4\\
  \toprule
  SSIM~\cite{wang2004image}  
  & 0.162  & 0.086  & 81.4\% & 37.3\% & 0.073  & 0.074  & 97.4\% & 36.8\% \\
  FSIM~\cite{zhang2011fsim}  
  & 0.268  & 0.123  & 14.2\% & 50.0\% & 0.151  & 0.058  & 32.4\% & 37.5\% \\   
  VIF~\cite{sheikh2006image}   
  & 0.322  & 0.236  & 43.5\% & 44.1\% & 0.111  & 0.150  & 22.2\% & 52.8\% \\ 
  GMSD~\cite{xue2014gradient}  
  & 0.417  & 0.210  & 21.9\% & 42.6\% & 0.259  & 0.132  & 63.6\% & 58.6\% \\
  \toprule
  
  \textbf{Scoot(CE)}& \textbf{0.037}  & \textbf{0.025}  & \textbf{95.9\%} & \textbf{76.3\% }& \textbf{0.012}  & \textbf{0.008}  & \textbf{97.5\%} & \textbf{78.8\%} \\   
  \bottomrule
  \end{tabular}
\end{table*}

\subsection{Alternative Features}\label{sec:otherFeature}
As described in \secref{sec:methodology}, sketches are quite close to textures.
There are many other texture \& edge-based features (\eg GLRLM~\cite{galloway1974texture},
Gabor~\cite{gabor1946theory}, Canny~\cite{canny1987computational}, Sobel~\cite{sobel1990isotropic}).
Here, we select the most wide-used features as candidate alternatives to replace our ``CE'' feature.
For GLRLM, we select all five statistics mentioned in the original version.
Results are shown in \tabref{table:AlternativeTexture}.
Gabor and GLRLM are texture features, while the other two are edge-based. With all the texture
features our measure provides a high consistency with human ranking (MM4). However,
GLRLM does perform well according to MM1 \& 2 \& 3. Gabor is reasonable in terms of MM1 \& 2, but fails in MM3.
For edge-based features, Canny fails according to all meta-measures. Sobel is very stable to slight resizing (MM1) or rotating (MM2),
but fails to capture content (MM3) and is not consistent with human judgment (MM4). 

\begin{figure}[t!]
\centering
    \begin{overpic}[width=\columnwidth]{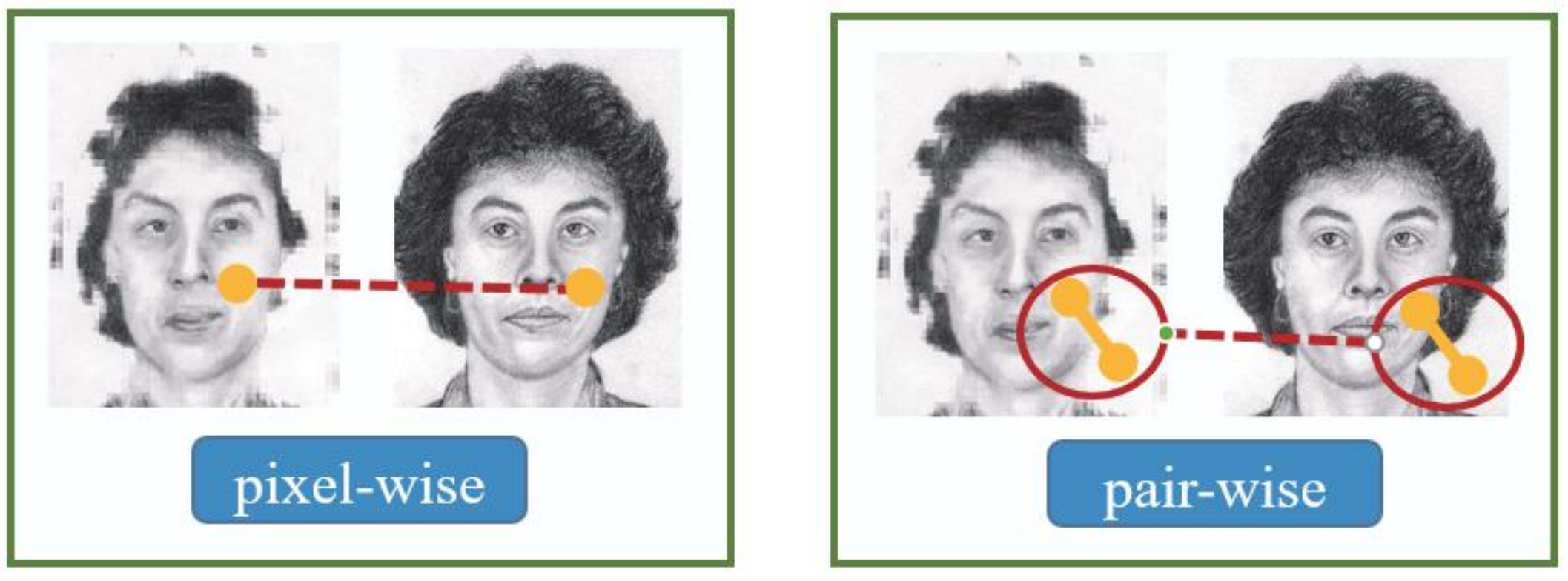}
    \end{overpic}
    \caption{A toy example illustrating the difference between our pair-wise measure (right) 
    and a pixel-wise measure (left).}\label{fig:Difference}
\end{figure}

\textbf{New insight.} Note that the alternative features \textbf{\textit{do not 
consider the pair-wise ``co-occurrence'' texture properties}} (see \figref{fig:Difference}), 
only our measure considers
this cue and provides the best performance compared to the other features. 
\tabref{table:AlternativeTexture} indicates that the pair-wise co-occurrence texture is suitable for FSS evaluation.

\begin{table}[t]
  \centering
  \footnotesize
  \renewcommand{\tabcolsep}{1.2mm}
  \caption{Quantitative comparison of different evaluation measures on our 4 meta-measures.
  MM: meta-measure. Scoot(CE)$^\dag$ indicates the non-quantized setting.
  The top measure for each MM are shown in boldface.
  These differences are all statistically significant at the $\alpha< 0.05$ level.}\label{table:AlternativeTexture}
  \begin{tabular}{l|cccc|cccc}
  \toprule
  &\multicolumn{4}{c|}{CUFS} & \multicolumn{4}{c}{CUFSF}\\
  \cmidrule{2-9}
  Measure & MM1 & MM2 & MM3 & MM4 & MM1 & MM2 & MM3 & MM4\\
  \toprule
  Scoot(Canny)& 0.086  & 0.078  & 33.7\% & 27.8\% & 0.138  & 0.146  & 0.0\%  & 0.1\% \\    
  Scoot(Sobel) & 0.040  & 0.037  & 0.0\% & 32.8\% & 0.048  & 0.044  & 0.0\% & 52.6\% \\
  
  \toprule
  Scoot(GLRLM)& 0.111  & 0.111  & 18.6\% & 73.7\% & 0.125  & 0.079  & 64.6\% & 68.0\% \\ 
  Scoot(Gabor)& 0.062  & 0.055  & 0.0\%  & 72.2\% & 0.089  & 0.043  & 19.3\% & \textbf{80.9\%} \\  
  \toprule
  Scoot(CE)$^\dag$ & \textbf{0.022}  & \textbf{0.014}  & 48.8\% & 73.1\% & 0.012  & \textbf{0.007}  & 92.5\% & 79.4\% \\   
  \textbf{Scoot(CE)}& 0.037  & 0.025  & \textbf{95.9\%} & \textbf{76.3\% }& \textbf{0.012}  & 0.008  & \textbf{97.5\%} & 78.8\% \\   
  \bottomrule
  \end{tabular}
\end{table}

\section{Conclusion}
In this paper, we summarized the widely-used image quality measures for
face sketch synthesis evaluation based on ``pixel-level'' degradation,
and enumerated their limitations. Then we focused on the
basic principles of sketching and proposed the use of structure
co-occurrence texture as an alternative motivating principle for
the design of face sketch style similarity measures.
To amend the flaws of existing measures, we proposed
the novel \textbf{Scoot-measure} which simultaneously captures both ``block-level'' spatial structure and co-occurrence
texture. Also, we concluded that the pair-wise co-occurrence properties 
are more important 
than pixel-level properties for style similarity evaluation of face sketch synthesis images.

In addition, extensive experiments with 4 new meta-measures
show that the proposed measure provides a reliable evaluation and achieves
the best performance.
Finally, we created two new human ranked
face sketch datasets (containing 776 images \& 1888 images) to examine the correlation between evaluation
measures and human judgment. The study indicated a higher degree (78.8\%)
of correlation between our measure and human judgment.
The proposed Scoot-measure is motivated from substantially
different design principles, and we see it as complementary to the existing
approaches.

All in all, our measure provides new insights into face sketch synthesis
evaluation where current measures fail to truly examine the pros and cons of 
synthesis algorithms. We encourage the community to consider this measure in 
future model evaluations and comparisons.

\textbf{Future works.}
Though our measure is only tested on face sketch datasets, it is not limited
to evaluating face sketch synthesis. 
We will also investigate the potential to use the measure to describe styles in
sketches and investigate new face sketch synthesis models.
To promote the development of this field, our code and dataset will be made
publicly available.


%
%

\bibliographystyle{ACM-Reference-Format}
\bibliography{SketchMeasure-bibliography}

\end{document}